\documentclass[letterpaper]{article} 
\usepackage{aaai2027}  
\usepackage[hyphens]{url}  
\usepackage{graphicx} 
\usepackage{natbib}  
\usepackage{caption} 
\usepackage{amsmath} 
\usepackage{amssymb} 

\usepackage{booktabs}

\setcitestyle{maxnames=6,minnames=1}

\nocopyright 

\title{PTC-Decoder: Towards Intelligent SLMs on Offline Resource-Constrained Edge Devices}
\author {
    Minghui Yu\textsuperscript{\rm 1},
    Ke Mu\textsuperscript{\rm 2},
    Gang Wu\textsuperscript{\rm 1}\corresponding
}
\affiliations {
    \textsuperscript{\rm 1}Shanghai Jiao Tong University, Computer Institute\\
    \textsuperscript{\rm 2}Shanghai Landfun Information Technology\\
    yminghui@sjtu.edu.cn, muke2433@gmail.com, dr.wugang@sjtu.edu.cn
}

\begin{document}

\maketitle

\begin{abstract}
  Deploying small language models (SLMs) on offline, resource-constrained edge devices such as remote sensing satellites presents a fundamental challenge: their limited reasoning capacity hinders reliable execution of multi-step agent tasks requiring complex tool orchestration. Existing plan-solve paradigms rely on prompt-based enforcement, which our experiments show SLMs almost entirely disregard: weak models fail to invoke the plan. We propose PTC‑Decoder (Plan‑Tool Constrained Decoder), a training‑free, plug‑and‑play decoder framework that combines (1) a Plan‑to‑Act paradigm, which elevates planning to an atomic tool and forces its invocation at the first inference step, and (2) TC‑Decoder, a deterministic finite automaton that imposes token‑level hard constraints on tool names while preserving freedom over parameter generation, thereby retaining SLM reasoning capability. Evaluated on 200 real remote‑sensing satellite tasks across 7 SLMs, PTC‑Decoder yields a statistically significant mean overall score gain of +1.21 (\(p<0.01\), 95\% CI [+1.13, +1.29]), with consistent improvements across models and other datasets. An ablation study that removes TC-Decoder causes substantial performance degradation across all quality metrics without reducing computational cost, confirming TC‑Decoder as the primary driver. PTC‑Decoder thus offers a lightweight yet effective solution for improving step‑level reliability, with final‑answer accuracy remaining an open challenge. In essence, we enforce plan adherence by constraining the permissible output vocabulary during inference, without requiring retraining.
\end{abstract}

\begin{links}
    \link{Code}{https://github.com/yuminghui/llm-tool-constrained-decoder}
\end{links}

\section{Introduction}
\begin{figure}[t]
    \centering
    \includegraphics[width=\columnwidth]{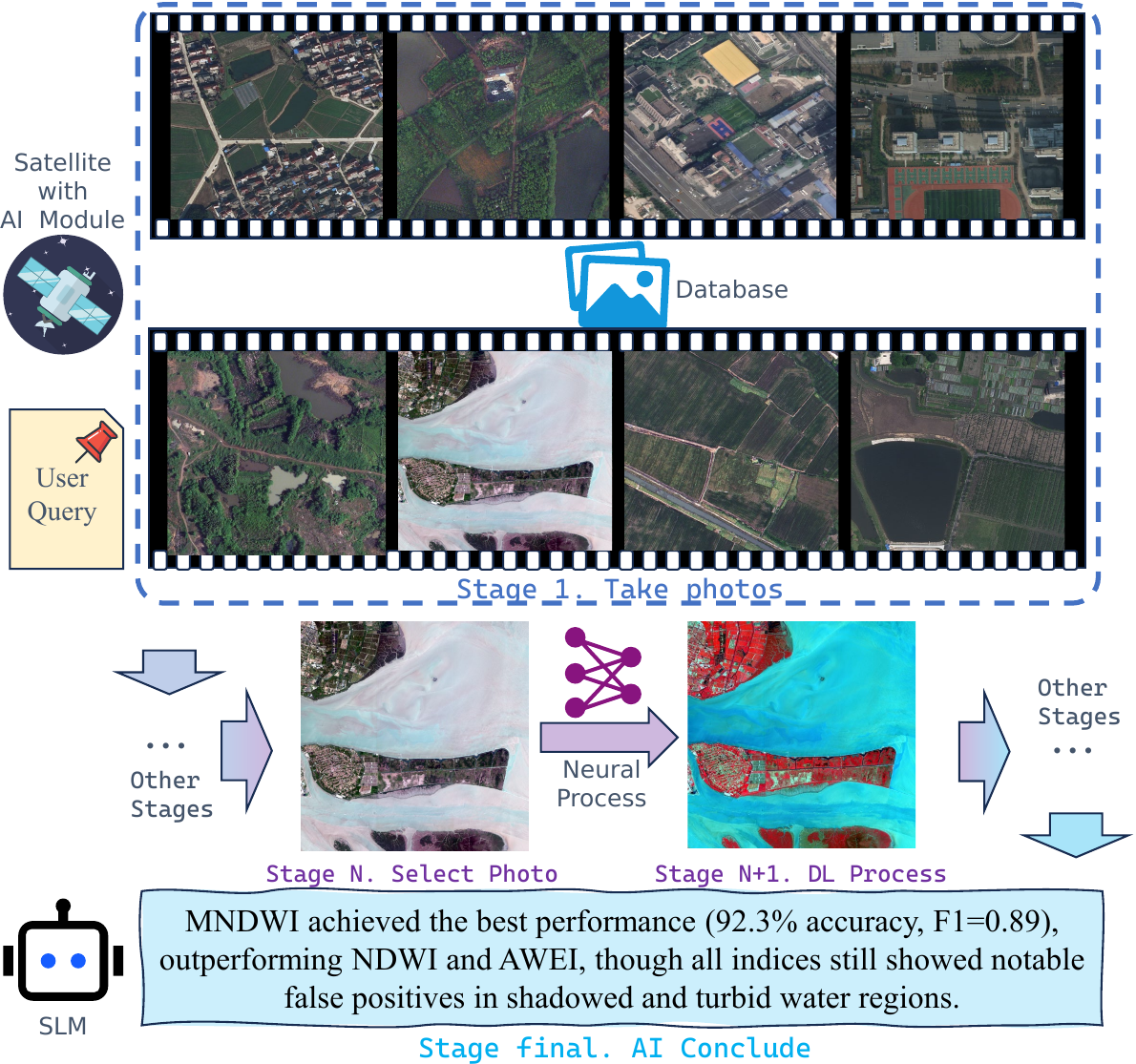}
    \caption{Illustration of the in‑orbit task processing flow for a satellite with an onboard agent.}
    \label{fig:complete_task_demo}
\end{figure}

Large language models (LLMs) have achieved remarkable success and continuous evolution across diverse scenarios, including software engineering \cite{10.1145/3695993}, intelligent customer service \cite{hong-etal-2025-augmenting}, and enterprise knowledge retrieval \cite{mishra2026systematicframeworkenterpriseknowledge,zhao2026surveylargelanguagemodels}. However, deploying LLMs on offline edge devices is challenging due to severe resource constraints (e.g., unified memory $\leq$ 16 GB) and competing resource-intensive applications. In our satellite Agent system (Figure~\ref{fig:complete_task_demo}) running on an NVIDIA Jetson Orin NX, high-resolution image recognition, remote sensing index analysis, and change detection leave less than 4 GB of memory for SLM deployment. We address this by constraining the model's output vocabulary during decoding, forcing tool adherence without training.

Before presenting our method, we first examine why existing studies fall short in this setting. They primarily pursue two approaches to enhance SLM-based agent capabilities: (1) end-to-end training \cite{erdogan-etal-2024-tinyagent} and knowledge distillation \cite{chen-etal-2025-drag,sharma2025smalllanguagemodelsagentic}, (2) plan-then-execute paradigm \cite{wang-etal-2023-plan}. However, the former relies on training data, which is extremely scarce in the remote sensing agent domain, rendering such methods inapplicable to our application domain. The latter, although significantly boosting LLM agent performance, is found by our experiments (Table \ref{tab:prompt-only}) to be incapable of ensuring stable planning in SLMs through prompt constraints alone.

In resource-constrained edge device scenarios, SLMs are typically adopted. Despite the rapid development of LLMs bringing training optimizations (e.g., GRPO \cite{deepseek-math}) and architectural improvements (e.g., sliding window attention \cite{yu2026swaaslidingwindowattention,gemmateam2025gemma3technicalreport}, gated linear attention) that enhance SLMs' reasoning capabilities and context length \cite{lu-etal-2025-regla}, they still suffer from severe hallucinations, insufficient reasoning ability, and misuse of tools when handling complex agent tasks and long contexts \cite{sun2025onionevalunifiedevaluationfactconflicting}, making it difficult to reliably execute.

\begin{table}[t]
\centering

\begin{tabular}{@{}l rr|rr@{}}
\toprule
& \multicolumn{2}{c|}{Baseline}
& \multicolumn{2}{c}{Prompt-Plan} \\
\cmidrule(lr){2-3} \cmidrule(lr){4-5}
Model
& P.R. & Suc.
& P.R. & Suc. \\
\midrule
Qwen3-1.7B   & .10 & 1.00 & .88 & .71 \\
Qwen3.5-2B   & .00 & .79  & .13 & .86 \\
Qwen3.5-0.8B & .07 & .92  & .20 & .99 \\
Qwen3-0.6B   & .03 & .99  & .47 & 1.00 \\
Gemma-4-2B(Q4) & .00 & 1.00 & .66 & 1.00 \\
Gemma-3-1B   & .00 & 1.00 & .00 & 1.00 \\
DeepSeek-R1-1.5B & .00 & 1.00 & .00 & 1.00 \\
\bottomrule
\end{tabular}
\caption{%
  Prompt-Plan enforcement versus baseline.
  The former adds \textit{MUST call plan first} into system prompt instructing the model to call $plan$.
  P.R.:~plan call rate; 
  Suc.:~success rate, higher the value, fewer times the agent crashes.
  Weaker models completely ignore the prompt (P.R.$=$0),
  and even capable models show degraded performance.
}
\label{tab:prompt-only}
\end{table}

Furthermore, some studies adopt task decomposition and edge-cloud collaboration \cite{ecoagent}, where simple tasks are processed locally while complex ones are offloaded to the cloud \cite{hybridslmandllmforedgecloud,li2025collaborativeinferencelearningedge}, offering a new perspective for enhancing edge intelligence. However, most remote sensing satellites operate as offline devices without Internet access \cite{wu2025nearrealtimeearthobservationstarlink} and rely solely on microwave or laser links for ground communication \cite{datadelayandoverflowattacksinearthobs}, making the powerful cloud-based LLMs inaccessible to our Agent. Current SLM research still fails to address the instruction‑following deficiency of SLMs in offline resource-constrained scenarios.

To this end, we propose \textbf{P}lan-\textbf{T}ool \textbf{C}onstrained \textbf{Decoder} (\textbf{PTC-Decoder}), a decoder framework that enforces planning and tool invocation at the decoding level. The initial attempt—forcing the agent to generate a plan at the first step via prompting—proved ineffective (Table~\ref{tab:prompt-only}). Meanwhile, the existing constrained decoders cannot be directly restricted from a specific tool level \cite{constraineddecoder}. We thus introduce TC-Decoder, which intervenes in the decoding process to deterministically enforce tool-name outputs at designated steps (while leaving parameters to the model). Integrating both—planning first, then constrained tool execution—yields PTC-Decoder. Although this may reduce model creativity, experiments show that appropriate constraints better utilize SLMs' limited reasoning for complex tasks. The decoder is plug-and-play, requiring no training.

Overall, our contributions are threefold:
\begin{enumerate}
    \item We propose \textbf{PTC‑Decoder}, a plug‑and‑play decoder with Plan‑to‑Act tailored for SLM Agents. This design mitigates the reasoning bottleneck on offline edge devices and significantly improves performance on a satellite agent benchmark and other datasets.
    \item We construct and open-source a compact yet high-quality on-board Agent operational benchmark, providing a foundational reference and starting point for the future construction of more comprehensive databases.
    \item Across multiple benchmark metrics, PTC-Decoder significantly outperforms the baseline, achieving a mean overall gain of +1.21 across 7 SLMs, with the weakest model improving by 350\% and capable models reaching an F1 of up to 0.359 against ground-truth tool sequences.
\end{enumerate}

Our SLM agent investigated in this study deploys on commercial satellites (16GB) to support ground control with remote sensing services.
\begin{figure*}[t]
    \centering
    \includegraphics[width=\textwidth]{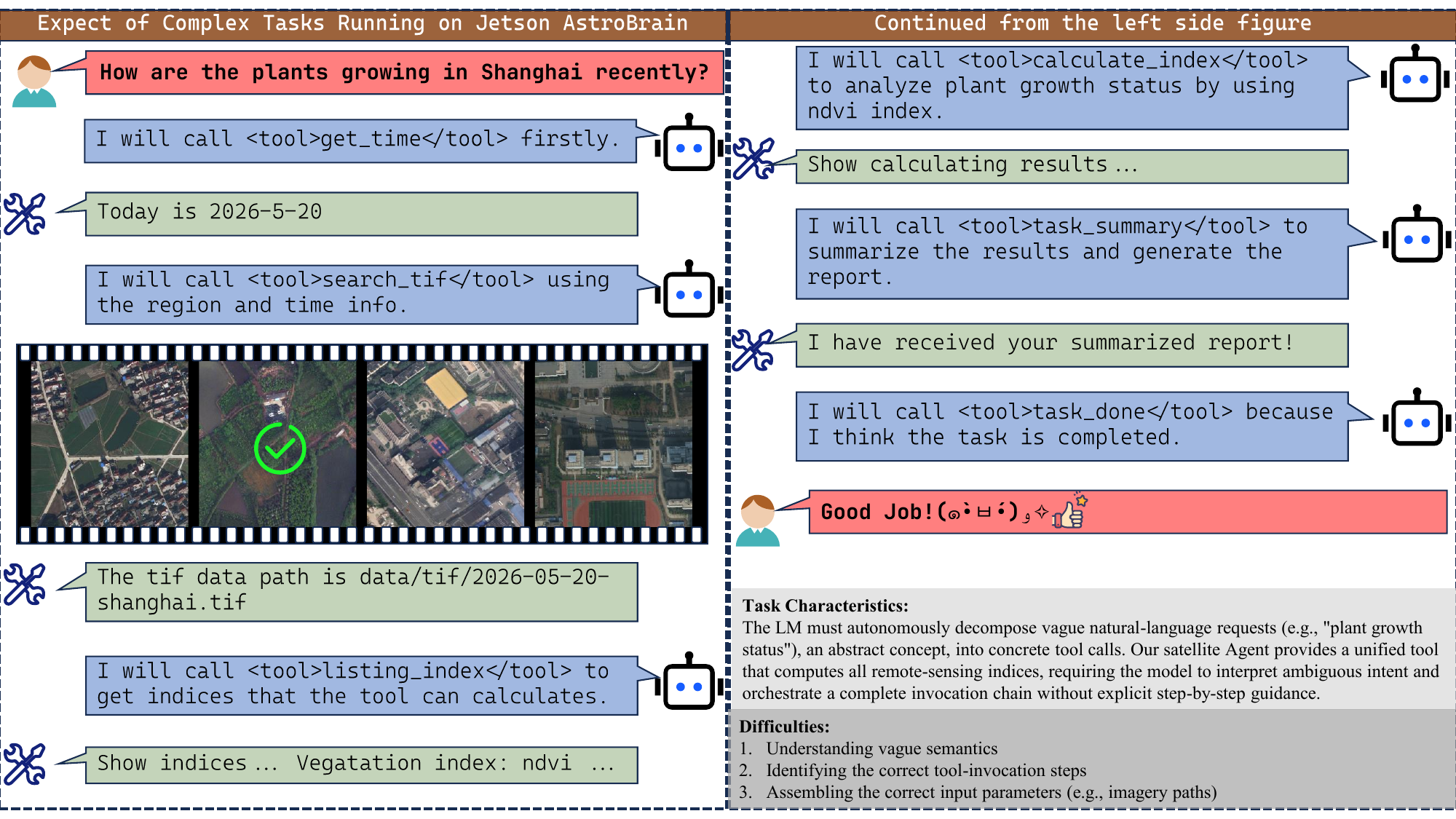}
    \caption{Example of the Expected Complex Task Workflow in the Remote Sensing Satellite Agent System.}
    \label{fig:complex_task_example}
\end{figure*}

\section{Related Work}
\subsection{Application of SLMs on Edge Devices}
As large language model capabilities continue to advance \cite{attention_is_all_you_need,gpt3techreport,CoT}, deploying them onto resource-constrained edge devices like smartphones and IoT terminals has become a key focus in both academia and industry. An empirical study covering over 60 SLMs for edge deployment reveals that \cite{lu-etal-2025-demystifying}, despite growing intelligence across tasks, substantial room remains for optimizing in-context learning and operational efficiency \cite{ni2026following}. Hardware acceleration, edge-cloud collaboration \cite{11075838}, model optimization, and deployment optimization constitute the main research thrusts for SLM edge adoption \cite{10.1145/3719664}. In addition, limited passive thermal dissipation capacity makes thermal throttling from continuous inference a major challenge \cite{WANG2025100755}.

\subsection{Lightweight Architectures and Model Quantization}
The core challenge of deploying language models on edge devices lies in balancing model size, inference speed, and task performance. Lightweight Transformer architectures and model quantization are two primary approaches \cite{samson2026lightweighttransformerarchitecturesedge,zhou-etal-2025-revisiting}; quantization can compress model size by 4–10× while retaining 75\%–96\% accuracy \cite{zhou-etal-2025-revisiting}. Meanwhile, algorithms such as MobiLoRA focus on optimizing LoRA-adapted LLM inference on mobile devices, particularly for KV Cache storage constraints \cite{li-etal-2025-mobilora}.

\subsection{Edge-Cloud Collaboration}
Beyond on-device deployment, edge-cloud collaboration and multi-LLM cooperation have emerged as active research directions \cite{11075838,zhu2025celslmefficientlargesmalllanguage,multiagent1,multiagent2,multiagent3}. These architectures offload complex tasks to the cloud while keeping simple ones local, effectively alleviating compute bottlenecks, reducing on-device cost, and partially enhancing SLM intelligence on edge devices \cite{zhu2025celslmefficientlargesmalllanguage}. However, such solutions are entirely inapplicable for offline devices without network access such as our remote-sensing satellite. How to compensate for SLM intelligence deficiencies on non-networked edge devices remains a largely unexplored pain point, and this work specifically targets this challenge.

\section{Methodology}
\begin{figure*}[t]
    \centering
    \includegraphics[width=\textwidth]{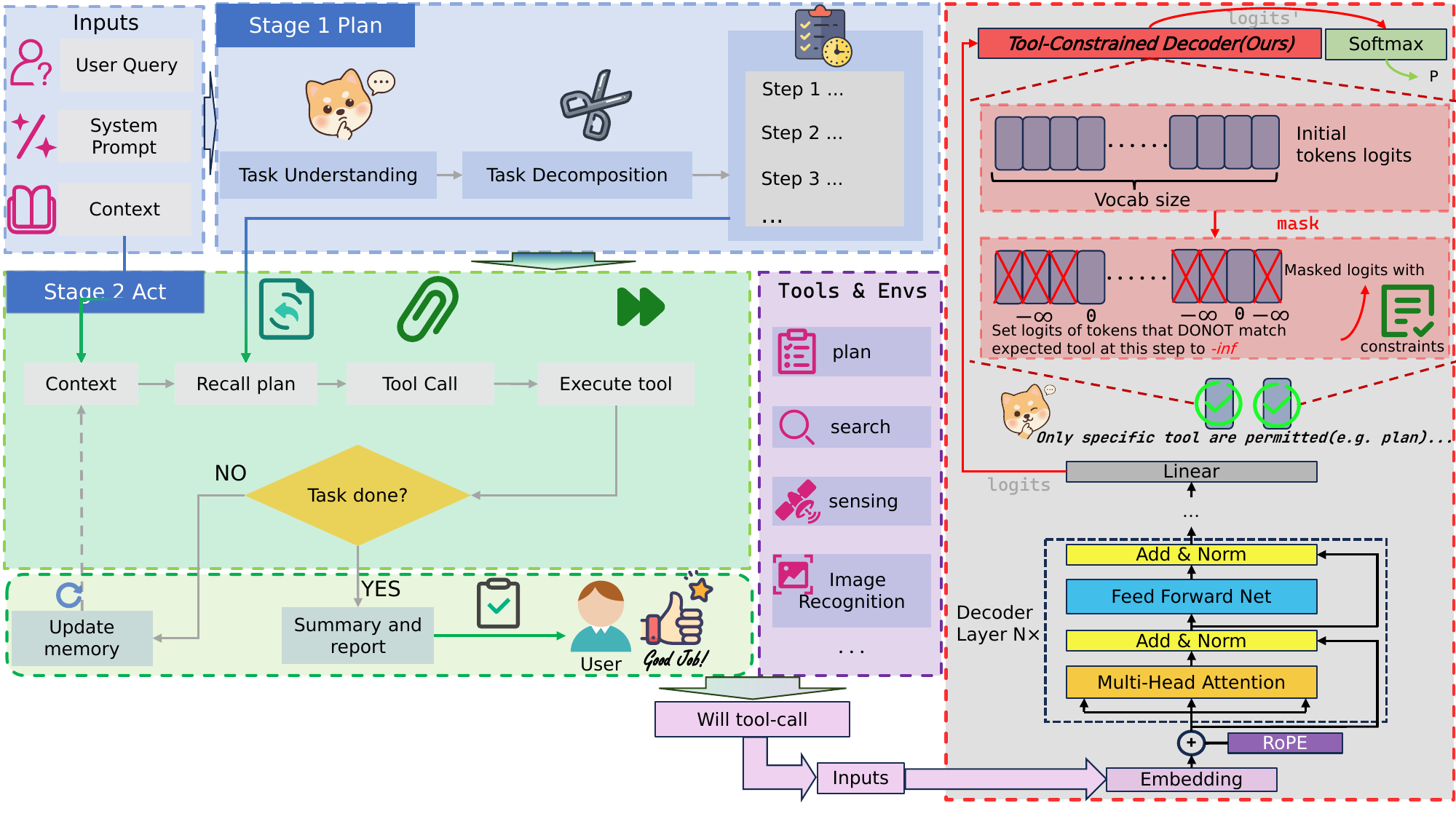}
    \caption{Our complete PTC-Decoder architecture}
    \label{fig:ptcdecoder}
\end{figure*}

We aim to enhance SLM step-level tool-calling adherence for edge agent tasks. This is critical, as we need to perform highly complex remote-sensing tasks—such as “How are the plants growing in Shanghai recently?”—with the expected workflow shown in Figure~\ref{fig:complex_task_example} \cite{NEURIPSDATASETSANDBENCHMARKS2021_LOVEDA}. However, models deployable on our satellite are typically quantized small models ($\leq$2B), for which such complex tasks are excessively difficult, rendering correct end-to-end agent execution nearly infeasible. To overcome this challenge, we design \textbf{PTC-Decoder}, as shown in Figure \ref{fig:ptcdecoder}.

This architecture prevents the model from invoking unexpected tools via step-by-step constraints, while still allowing autonomous parameter determination, keeping reasoning within a reasonable boundary, and preventing rapid context expansion that would otherwise cause the SLM to become lost. 

\subsection{Plan-to-Act}
To prevent the SLM agent from becoming lost in tool-invocation contexts when facing ambiguous complex tasks, our PTC-Decoder adopts a "Plan-to-Act" decision paradigm (left part in Figure~\ref{fig:ptcdecoder}). Its core idea is that in each interaction round, the agent first generates a high level plan and then strictly follows this plan to invoke atomic tools step by step. This design is particularly well-suited for resource-constrained edge deployment scenarios.

\noindent \textbf{Motivation: Resource Bottlenecks in Edge Computing.} Our SLM is deployed on a small remote-sensing satellite with only 16GB of unified memory, of which less than 4GB is available for model inference. Under such extreme constraints, directly confronting the agent with ambiguous, multi-step, context-dependent user requests leads to cognitive overload as the model attempts to simultaneously handle task decomposition, tool selection, parameter filling, and result interpretation, manifesting as invalid tool calls, missing critical steps, repetitive loops, or unparseable outputs. LLM agent paradigms such as ReAct and Plan-and-Solve rely on hundred-billion-parameter models and multiple inference passes for robust planning \cite{yao2023react,wang-etal-2023-plan}, which is infeasible on resource-constrained satellite platforms.

\noindent \textbf{Implementation: Planning as a Tool.} We define "plan" as a tool within the agent system: its input is a natural-language task description, and its output is a structured sequence of atomic tool names. The key design decision is that the first inference round must forcibly invoke the `plan` tool—rather than allowing free selection—which is achieved via the tool-constrained decoder detailed in the next section. After the plan returns a sequence, each subsequent agent loop invokes the corresponding atomic tools in order. Formally, letting the full toolset be $\mathcal{T}=\{\text{plan}, t_2, t_3, \dots, t_n\}$ and the user task be $\mathcal{P}$, $\mathcal{C}$ be the request initiated by the SLM, and $\mathcal{S}$ the corresponding tool call schema, the first-round output $x^{(1)}$ must satisfy:
\begin{equation}
    \forall \mathcal{P}, ~ x^{(1)} \in \mathcal{C}_{plan} \overset{\triangle}{=} \{\mathcal{S} ~|~ \mathcal{S}.name==plan\}
\label{eq:reasoning_first_call}
\end{equation}

\subsection{Tool-Constrained Decoder}
Like LLMs, SLMs are modeled as causal language models, performing autoregressive sampling by continuously predicting the next token \cite{attention_is_all_you_need}. 
Modern decoder-only LLMs rely on autoregressive sampling from the model's own probability distribution at each step, granting them strong open-domain text generation capabilities \cite{gpt3techreport}. However, for SLMs deployed on offline edge devices, this unconstrained mechanism often fails—weak models may disregard prompt instructions for tool invocation, severely undermining their reliability and practicality in complex agent tasks.

To address this, we propose the \textbf{Tool-Constrained Decoder}, which forcibly injects the desired tool during sampling, ensuring that the SLM generates exactly that tool invocation at the designated step. This module is lightweight and plug-and-play: it requires no changes to existing agent implementations, no modifications to stable prompts, and no additional training. Compared to SFT or RL approaches, our method is far less intrusive.

Different from a general constraint decoder, this design avoids over‑constraint, preserving parameter generation quality and reasoning capability. The module is plug‑and‑play: it can be disabled for unconstrained steps and enabled on demand. The right part in Figure~\ref{fig:ptcdecoder} illustrates the decoder structure with the module activated.

Let \( \mathcal{V} \) denote the vocabulary, \( x_{<t} \) the generated sequence up to step \( t-1 \), \( l_t \in \mathbb{R}^{|\mathcal{V}|} \) the raw logits at step \( t \), \( s_t \) the internal state of the constrained decoder at step \( t \) (a finite-state machine characterizing the preconditions satisfied by the current sequence)

Let the original conditional distribution $\pi_t$ of the language model be:
\begin{equation}
    \pi_t(v) \overset{\triangle}{=} \textbf{P}_\mathcal{LM}(x_t=v|x_{<t}) = \text{softmax}(l_t)[v], ~ v\in \mathcal{V}
\label{eq:initial_probs}
\end{equation}
A generic constrained decoder can be formally defined as applying a mask function $M_t$ to the raw logits via Hadamard product to modify the distribution before output of the final logits at time step $t$ \cite{constraineddecoder}:
\begin{equation}
    \hat{l_t} = l_t \odot M_t
\label{eq:initial_constrained}
\end{equation}
where $\hat{l_t}$ represents the logits modified by the mask, which corrects the raw scores of all tokens violating the rules to $-\infty$ so that their corresponding token probabilities approach 0 infinitely after the softmax operation; the final corrected token probability distribution $\hat{\pi}_t(v)$ is given by:
\begin{equation}
    \hat{\pi}_t(v) \overset{\triangle}{=} \textbf{P}_\mathcal{LM}(x_t=v|x_{<t}) = \text{softmax}(\hat{l_t})[v], ~ v\in \mathcal{V}
\label{eq:after_constrained_probs}
\end{equation}

We design an agent-oriented constrained decoder (Tool-Constrained Decoder). Unlike approaches that constrain the entire output space, our decoder imposes token-level hard constraints only on tool invocation format and tool names, leaving parameter content entirely to the underlying SLM. This design aligns with the plan-to-act paradigm: we enforce the use of designated tools, but delegate the generation of action details to the model's own knowledge and contextual reasoning.

To implement this partial constraint, we construct a prefix-constrained automaton $\mathcal{A}_{tool}(Q,\Sigma,\delta,q_0,F)$ based on a Deterministic Finite Automaton (DFA), where:
\begin{itemize}
\item $Q=Q_{fix}\cup \{q_{free}\}$, where $Q_{fix}$ is the state sequence for the fixed prefixes, and $q_{free}$ is the state sequence for subsequent unconstrained tokens.
\item $\Sigma$ is the set of all tokens in the vocabulary $\mathcal{V}$.
\item On $Q_{fix}$, the transition function $\delta$ forces the generation of preset tokens according to specified rules, such as \texttt{<|name>get\_weather<name|>} among \texttt{<|tool><|name>get\_weather<name|> <|arguments>...<arguments|><tool|>}; once the fixed sequence generation is complete, the automaton transitions from $Q_{fix}$ to the absorbing state $q_{free}$ via any token, and $q_{free}$ has self-loop transitions for any token in $\Sigma$, i.e., $\forall v\in \mathcal{V},\delta(q_{free},v)=q_{free}$.
\item $F\in q_{free}$, where the terminal state is reached through the EOS token, i.e., $\delta(q_{free}, v_{eos})=F$.
\end{itemize}

\begin{table*}[t]
\centering
\begin{tabular}{@{}l c|cccc c|cccc c|cccc@{}}
\toprule
& \multicolumn{5}{c}{Baseline}
& \multicolumn{5}{c}{PTC-Decoder (Ours)}
& \multicolumn{5}{c}{Plan w/o TC-Decoder} \\
\cmidrule(lr){2-6} \cmidrule(lr){7-11} \cmidrule(lr){12-16}
Model & H. & Ov. & R.A. & F.R. & Rb. & H. & Ov. & R.A. & F.R. & Rb. & H. & Ov. & R.A. & F.R. & Rb. \\
\midrule
Qwen3-1.7B      & 2.05 & 1.98 & 0.87 & 2.12 & 2.71  & 3.10 & 2.94 & 2.32 & 2.97 & \textbf{3.06}  & 2.45 & 2.67 & 1.34 & 2.94 & 3.03 \\
Qwen3.5-2B      & 1.70 & 1.28 & 0.41 & 1.19 & 2.25  & 3.35 & \textbf{3.12} & \textbf{2.81} & 3.27 & 2.73  & 1.50 & 1.34 & 0.30 & 1.41 & 2.22 \\
Qwen3.5-0.8B    & 1.05 & 1.15 & 0.42 & 0.96 & 2.04  & 2.55 & 2.29 & 1.73 & 2.31 & 2.67  & 1.55 & 1.51 & 0.53 & 1.64 & 2.17 \\
Qwen3-0.6B      & 0.95 & 1.27 & 0.74 & 1.15 & 2.02  & 2.40 & 2.23 & 1.60 & 2.23 & 2.44  & 1.10 & 1.27 & 0.23 & 1.05 & 2.63 \\
Gemma-4-2B(Q4)  & 1.90 & 1.71 & 0.95 & 1.69 & 2.55  & 3.25 & 3.05 & 2.42 & \textbf{3.31} & 2.60  & 1.55 & 1.36 & 0.22 & 1.45 & 2.64 \\
Gemma-3-1B      & 0.40 & 0.73 & 0.33 & 0.43 & 1.84  & 1.70 & 1.56 & 0.94 & 1.48 & 2.25  & 0.90 & 1.00 & 0.14 & 0.87 & 2.10 \\
DeepSeek-R1-1.5B & 0.20 & 0.40 & 0.20 & 0.12 & 1.40 & 1.75 & 1.80 & 0.97 & 1.56 & 2.73  & 0.85 & 0.93 & 0.11 & 0.74 & 2.08 \\
\bottomrule
\end{tabular}
\caption{%
  LLM-judge scores across experimental conditions, with a Human double-check
  (H.) for each group (20 randomly sampled tasks per group, rated on the
  same $[0,5]$ by two annotators).
  The best LLM-judge value for each metric is \textbf{bolded}.
}
\label{tab:main-judge}
\end{table*}

Based on the aforementioned automaton $\mathcal{A}_{tool}$, we define the mask function $M_t^{'}(v)$ at the time step $t$ as:
\begin{equation}
M_{\text{t}}^{'}(v) = 
\begin{cases}
0, & \text{if } \delta(q_t, v) \text{ is defined},\\
-\operatorname{sgn}(l_t[v]) \cdot \infty, & \text{otherwise}.
\end{cases}
\label{eq:tool_constrained_mask}
\end{equation}
where $q_t$ is the state of the automaton at time step $t$, and the modified logits and probability distribution formally still follow:
\begin{equation}
    \hat{l_t^{'}} = l_t \odot M_t^{'}
\label{eq:my_constrained}
\end{equation}
\begin{equation}
    \hat{\pi}^{'}_t(v) \overset{\triangle}{=} \textbf{P}^{'}_\mathcal{LM}(x_t=v|x_{<t}) = \text{softmax}(\hat{l_t^{'}})[v], ~ v\in \mathcal{V}
\label{eq:my_constrained_probs}
\end{equation}

\section{Experiments}
\subsection{Experimental Setup}

\begin{table*}[t]
\centering
\begin{tabular}{@{}l cccr|cccr|cccr@{}}
\toprule
& \multicolumn{4}{c|}{Baseline}
& \multicolumn{4}{c|}{PTC-Decoder (Ours)}
& \multicolumn{4}{c}{Plan w/o TC-Decoder} \\
\cmidrule(lr){2-5} \cmidrule(lr){6-9} \cmidrule(lr){10-13}
Model & R. & P. & F1 & Tokens & R. & P. & F1 & Tokens & R. & P. & F1 & Tokens \\
\midrule
Qwen3-1.7B & .187 & .413 & .249 & 3437+181 & .305 & .295 & .292 & 7018+399 & .252 & \textbf{.647} & .355 & 7018+370 \\
Qwen3.5-2B & .123 & .137 & .123 & 3424+185 & \textbf{.388} & .284 & .318 & 7229+646 & .147 & .214 & .162 & 7000+509 \\
Qwen3.5-0.8B & .104 & .180 & .125 & 3424+137 & .229 & .237 & .227 & 7000+425 & .171 & .325 & .212 & 7000+421 \\
Qwen3-0.6B & .148 & .167 & .155 & 3437+112 & .250 & .249 & .246 & 7018+356 & .000 & .000 & .000 & 7018+262 \\
Gemma-4-2B(Q4) & .246 & .328 & .279 & 3130+112 & .379 & .361 & \textbf{.359} & 6435+868 & .097 & .128 & .109 & 6435+422 \\
Gemma-3-1B & .000 & .000 & .000 & 631+513 & .063 & .095 & .076 & 1437+809 & .000 & .000 & .000 & 1437+1103 \\
DeepSeek-R1-1.5B & .000 & .000 & .000 & 612+427 & .074 & .109 & .088 & 1368+567 & .000 & .000 & .000 & 1368+674 \\
\bottomrule
\end{tabular}
\caption{%
  Recall, Precision, F1 (per-task average), and average tokens (input + output).
  The best value for each metric is \textbf{bolded}.
}
\label{tab:main-coverage}
\end{table*}

\noindent \textbf{Edge Device.} All experiments are conducted on an NVIDIA Jetson Orin NX Developer Kit, which shares the same configuration as the control computer of a small offline remote-sensing satellite: an Orin GPU, 8-core Cortex-A78AE CPU, 16GB unified memory, 877GB storage, running Ubuntu 22.04 LTS with CUDA 12.2. To closely replicate the extreme resource constraints of actual satellite operations, we concurrently run multiple remote-sensing algorithms and daemon processes alongside the Agent tasks on this device, thereby approximating real onboard resource-limited conditions.

\noindent \textbf{Small Language Models and Setup.}
To demonstrate the generalizability of this study, we select mainstream SLMs from the open-source community, with memory limitation that the total effective parameter count does not exceed 2B, and that they are released by well-known model providers. Ultimately, we identified a total of 7 SLMs \cite{gemmateam2025gemma3technicalreport,gemmateam2026gemma4technicalreport,yang2025qwen3technicalreport,Guo_2025deepseekr1}. All models use temperature 0.2, top-p 0.95, and a maximum of 10 turns per task (overflowed as a failure), with up to 512 new tokens per free-generation step and 256 per constrained tool call.

\noindent \textbf{Evaluation Datasets.}
To validate the optimization strategies for our remote-sensing satellite Agent system, we conducted on-board experimental evaluations. Given that spaceborne edge devices offer AI compute below 100 TOPS and suffer from thermal throttling under prolonged operation, we carefully controlled the evaluation dataset scale to ensure smooth comparative experiments across 7 models. To address the absence of dedicated remote-sensing Agent benchmarks, we curated 200 real user tasks from on-board operations and had them cross-annotated by three remote-sensing and computer vision experts, producing a test suite with expected Agent execution trajectories and ground-truth results. This dataset will be released along with the code to facilitate future research in this domain.

\noindent \textbf{Evaluation Methods and Metrics.}
Our evaluation combines LLM-judge scores (DeepSeek-V4-flash) \cite{deepseekai2026deepseekv4highlyefficientmilliontoken} with rule-based metrics.
LLM-judge includes Overall (Ov.), which integrates complete tracks and logs for scoring; Result Accuracy (R.A.); Flow Reasonableness (F.R.); and Robustness (Rb.), all rated on a $[0,5]$ scale.
Rule-based metrics include Recall (R.), Precision (P.), and F1-score, computed per task against benchmark ground-truth tool sets and averaged (excluded non-functional tools like \textit{task\_done}).
Due to thermal throttling from sustained inference on satellite onboard systems, each task is executed only once. Therefore, we group 200 tasks as paired units and employ the Wilcoxon signed‑rank test $p$ to assess the statistical significance of the observed improvements\cite{demsar2006statistical}. For LLM-Judge, we will randomly pick 20 results to do a manually double-check for Ov., marked as (H.). All scores are reported as means across 200 benchmark tasks.

\subsection{Overall Comparison}

\subsubsection{LLM-Judge Evaluation}

Baseline performance is limited: the strongest model (Qwen3‑1.7B) achieves Ov. 1.98, while weaker models score below 0.75 across all dimensions.

PTC‑Decoder consistently improves all models (mean Ov. +1.21, \(p<0.01\)), with gains inversely correlated to baseline capability—DeepSeek‑R1‑1.5B improves by 350\% (0.40$\rightarrow$1.80), while Qwen3‑1.7B gains 48\% (1.98$\rightarrow$2.94). Flow reasonableness shows the largest average gain (+1.35); Gemma‑4‑2B reaches the highest F.R. of 3.31, and Qwen3.5‑2B gains +2.08 to 3.27. Result accuracy improves by a mean of +1.27, with Qwen3.5‑2B achieving both the largest gain (+2.40, 0.41$\rightarrow$2.81) and the global best R.A. Robustness gains are moderate (mean +0.52), with Qwen3‑1.7B attaining the best Rb. of 3.06 under PTC‑Decoder. Qwen3.5‑2B obtains the highest Ov. of 3.12.

Plan w/o TC‑Decoder yields marginal Ov. improvement (mean +0.22), with Gemma‑4‑2B (–0.35) degrading. R.A. and F.R. follow similarly, with mean R.A. declining by 0.15. Rb. increases by a mean of +0.23, though Qwen3‑1.7B’s Rb. is slightly higher under PTC‑Decoder (3.06 vs. 3.03), contrasting with earlier observations where free execution gave the best Rb. Human double-check shows the same trends.

\subsubsection{Rule-Based Evaluation}

Table~\ref{tab:main-coverage} reports benchmark Recall, Precision, and F1. Baseline performance is limited (mean Recall 0.115, F1 0.133), with Gemma‑3‑1B and DeepSeek‑R1‑1.5B scoring zero across all metrics.

PTC‑Decoder consistently improves all three metrics: mean Recall +0.126 (Qwen3.5‑2B leading: 0.123$\rightarrow$0.388), Precision +0.058, and F1 +0.096 (Gemma‑4‑2B reaching global best F1=0.359). Weak models recover from zero to 0.063 and 0.074 Recall. Absolute F1 remains modest across conditions, reflecting the difficulty of full tool coverage, but gains are consistent across all models.

Plan w/o TC‑Decoder yields mixed results: mean Recall and F1 decline by 0.020 and 0.013, suggesting that unenforced planning may reduce ground‑truth coverage. Qwen3‑1.7B is the sole improver (F1 +0.106, Precision 0.647, global best), albeit with narrow Recall (0.252). Qwen3‑0.6B collapses to zero on all metrics. We specifically examined this set of zero-score cases. Under the plan-only condition without TC-Decoder, Qwen3‑0.6B simply treated the task as completed in all instances, which directly accounts for its zero scores across all metrics.

As Gemma‑3‑1B and DeepSeek‑R1‑1.5B are not tool‑invocation models, they cannot understand tools. Manual trace analysis confirms that, without TC‑Decoder, their R., P., and F1 on remote‑sensing tasks are zero (these metrics exclude non‑functional tools like \textit{task\_done}).

\begin{figure*}[t]
    \centering
    \includegraphics[width=\textwidth]{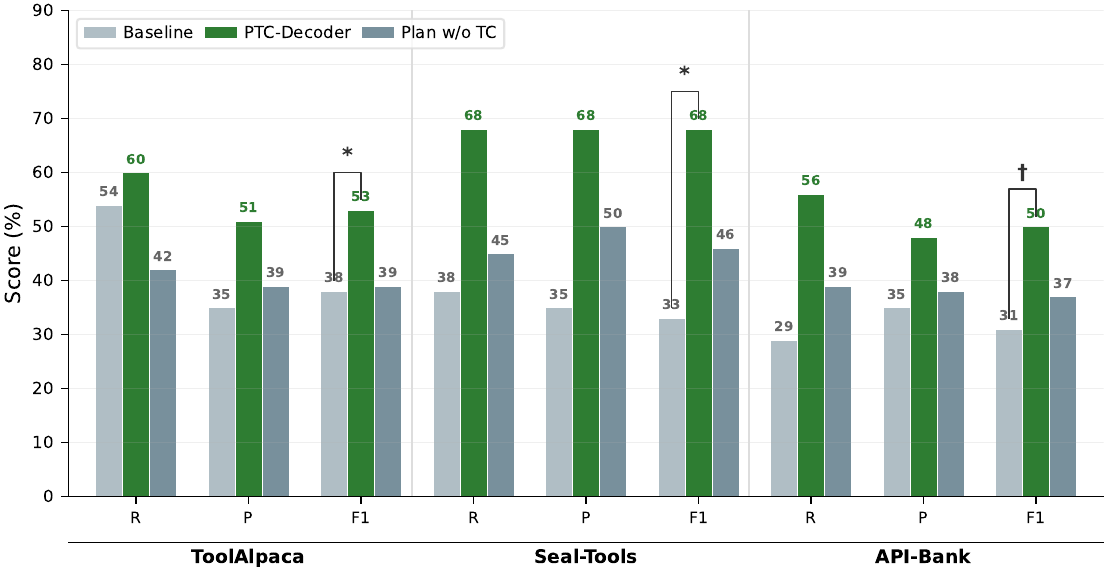}
    \caption{%
    Recall, Precision, and F1 across three external datasets (means over 7 SLMs).
    PTC-Decoder consistently improves all metrics.
    Plan w/o TC-Decoder shows inconsistent improvement, remaining close to baseline.
    Error bars omitted for clarity; within-group F1 SD ranges: ToolAlpaca 0.18--0.38, Seal-Tools 0.17--0.44, API-Bank 0.11--0.37.
    $*p<0.05$, $^\dagger p=0.06$ (Wilcoxon signed-rank test, $n=7$, PTC-Decoder vs.\ Baseline F1: ToolAlpaca $p=0.03$, Seal-Tools $p=0.03$, API-Bank $p=0.06$).
  }
    \label{fig:other_datasets}
\end{figure*}

\subsection{Performance on Other Datasets}

To test cross‑domain generalizability, we evaluate the same three experimental combinations on ToolAlpaca \cite{tang2023toolalpaca}, Seal‑Tools \cite{wu2025seal}, and API‑Bank \cite{li-etal-2023-api}, using the same 7 SLMs and experimental configs mentioned before. The overall results are shown in Figure~\ref{fig:other_datasets}. PTC‑Decoder consistently improves over baseline across all datasets: ToolAlpaca F1 38\% to 53\% (Recall 54\% to 60\%, Precision 35\% to 51\%); Seal‑Tools 33\% to 68\% (38\% to 68\%, 35\% to 68\%); API‑Bank 31\% to 50\% (29\% to 56\%, 35\% to 48\%). The largest gain appears on Seal‑Tools, where F1 more than doubles—driven by balanced improvements in both Recall and Precision, confirming that TC-Decoder enhances tool coverage while reducing off‑target invocations.

Plan w/o TC‑Decoder yields weaker results: ToolAlpaca F1 stagnates at 39\% (baseline 38\%), Seal‑Tools reaches 46\% (well below 68\%), and API‑Bank only 37\% (baseline 31\%). These results reinforce that plan‑level constraints alone are insufficient for most SLMs; execution‑level enforcement is the primary driver of the observed gains.

\subsection{Ablation Study}

To isolate TC-Decoder's contribution, we conduct an ablation study directly comparing PTC-Decoder against Plan w/o TC-Decoder, where the latter ablated TC-Decoder. Tables~\ref{tab:main-judge}~and~\ref{tab:main-coverage} report the full results.

\noindent \textbf{LLM-Judge metrics.}
Adding TC-Decoder improves Ov.\ by 0.99 on average across all 7 models (\(p<0.01\), bootstrap 95\% CI [+0.91, +1.06]), with Qwen3.5-2B (+1.78) and Gemma-4-2B (+1.69) gaining the most.
R.A. benefits substantially (mean +1.42): Qwen3.5-2B rises from 0.30 to 2.81 (\(\Delta=+2.51\)), and Gemma-4-2B from 0.22 to 2.42 (\(\Delta=+2.20\)).
F.R. improves by 1.00 on average; Gemma-4-2B reaches 3.31 (\(\Delta=+1.86\)), and Qwen3.5-2B 3.27 (\(\Delta=+1.86\)).
Rb. shows the smallest gain (mean +0.23), with Qwen3-1.7B's Rb.\ nearly unchanged (3.06 vs.\ 3.03), suggesting that TC-Decoder neither improves nor harms recovery behavior.

\noindent \textbf{Benchmark coverage.}
TC-Decoder consistently improves benchmark-aligned tool coverage.
Mean Recall rises from 0.095 to 0.241 (\(\Delta=+0.146\)), led by Gemma-4-2B (+0.282) and Qwen3.5-2B (+0.241, achieving the globally best Recall of 0.388).
F1 increases by 0.110 on average; Gemma-4-2B reaches the globally best F1 of 0.359 (+0.250 over its Plan w/o TC-Decoder counterpart).
The most pronounced effects are on models that collapse without TC-Decoder: Qwen3-0.6B's, Gemma-3-1B's, and DeepSeek-R1-1.5B's benchmark metrics all register zero under free execution, yet reach 0.246, 0.076, and 0.088 F1 with TC-Decoder.
Qwen3-1.7B is an interesting exception: its F1 is higher \textit{without} TC-Decoder (0.355 vs.\ 0.292), driven by Precision of 0.647 (globally best) under free execution, at the cost of Recall (0.252 vs.\ 0.305 with TC-Decoder).
This precision-recall trade-off illustrates that TC-Decoder's primary mechanism is enforcing broader tool coverage, which may modestly reduce precision for models already capable of focused execution.

\noindent \textbf{Efficiency.}
Token overhead is nearly identical between the two variants (2.07$\times$ vs.\ 2.08$\times$ over baseline), confirming that TC-Decoder adds negligible computational cost.

In summary, TC-Decoder is the decisive component of PTC-Decoder: it substantially improves LLM-judge scores and benchmark tool coverage across all models, transforms several SLMs from zero coverage to functional plan adherence, and does so without additional overhead.

\section{Conclusion}
This paper designed PTC-Decoder, a training-free, plug-and-play decoder framework that addresses the intelligence deficiency of SLMs on offline resource-constrained edge devices.
By coupling a Plan-to-Act paradigm with TC-Decoder, PTC-Decoder elevates 7 SLMs from weak baselines to functional agent performance, achieving a mean Ov.\ gain of +1.21 and peak F1 of 0.359.
Ablation confirms TC-Decoder as the decisive component. Removing it reduces F1 by 0.110 on average and degrades all quality metrics, yet offers no efficiency advantage.
PTC-Decoder thus provides a lightweight and effective solution for improving step-level reliability of SLM agents on offline edge devices.

\subsection{Limitations}
Despite improvements in all dimensions, R.A. remains the weakest dimension (best 2.81/5), and Rb. gains are modest (mean +0.52 vs. baseline). This suggests that TC-Decoder cannot guarantee correct answers or robust error recovery, since it only constrains tool names. Extending constraints to parameter schemas may help narrow this gap.
PTC-Decoder incurs a 2.07$\times$ token overhead. 
As a counterexample, Qwen3‑1.7B achieves higher F1 without TC‑Decoder, indicating that PTC‑Decoder primarily benefits intelligence‑deficient SLMs. As SLM capabilities advance, PTC-Decoder may become less necessary. For LLM, the creativity‑restricting effect of TC‑Decoder might outweigh its benefits.


\bibliography{references}


\end{document}